\documentclass{article}

\PassOptionsToPackage{numbers, compress}{natbib}

\usepackage[preprint]{neurips_2024}

\usepackage[utf8]{inputenc} 
\usepackage[T1]{fontenc}    
\usepackage{hyperref}       
\usepackage{url}            
\usepackage{booktabs}       
\usepackage{amsfonts}       
\usepackage{nicefrac}       
\usepackage{microtype}      
\usepackage{xcolor}         
\usepackage{graphicx}
\usepackage{subcaption}

\usepackage{amsmath}
\usepackage{amssymb}
\usepackage{mathtools}
\usepackage{amsthm}

\theoremstyle{plain}
\newtheorem{theorem}{Theorem}[section]

\theoremstyle{definition}

\theoremstyle{remark}

\usepackage{tikz}
\usepackage{bm}
\usepackage{enumitem}
\usepackage{wrapfig}

\newcommand{\R}{\mathbb{R}}

\newcommand{\E}{\mathbb{E}}
\newcommand{\T}{\top}
\newcommand{\eye}{\text{I}}

\DeclareMathOperator{\bigo}{\mathcal{O}}
\DeclareMathOperator{\diag}{diag}
\DeclareMathOperator{\tr}{Tr}
\DeclareMathOperator{\relu}{ReLU}
\newcommand{\kronecker}{\delta}

\definecolor{myblue}{HTML}{006BA4}
\definecolor{myorange}{HTML}{FF800E}

\title{Mean-Field Microcanonical Gradient Descent}

\author{
  Marcus Häggbom\thanks{Department of Mathematics, KTH Royal Institute of Technology} \\
  SEB Group \\
  Stockholm, Sweden \\
  \texttt{haggbo@kth.se} \\
  \And
  Morten Karlsmark \\
  SEB Group \\
  Stockholm, Sweden \\
  \texttt{morten.karlsmark@seb.se}
  \AND
  Joakim Andén \\
  Department of Mathematics \\
  KTH Royal Institute of Technology \\
  Stockholm, Sweden \\
  \texttt{janden@kth.se} \\
}

\begin{document}

\maketitle

\begin{abstract}
Microcanonical gradient descent is a sampling procedure for energy-based models allowing for efficient sampling of distributions in high dimension. 
It works by transporting samples from a high-entropy distribution, such as Gaussian white noise, to a low-energy region using gradient descent. 
We put this model in the framework of normalizing flows, showing how it can often overfit by losing an unnecessary amount of entropy in the descent. As a remedy, we propose a mean-field microcanonical gradient descent that samples several weakly coupled data points simultaneously, allowing for better control of the entropy loss while paying little in terms of likelihood fit. We study these models in the context of financial time series, illustrating the improvements on both synthetic and real data.
\end{abstract}

\section{Introduction}
\label{sec:intro}
The defining characteristic of a well-behaved generative model is the balance between its ability to, on the one hand, produce samples that are typical of the training data, while on the other hand having a significant amount of diversity within its samples. 
For example, a generative adversarial network (GAN) which has suffered mode collapse could produce great samples within one mode but not others.
Similarly, the empirical distribution of the training data approximates the training data well but is useless for generating new samples, while a Gaussian white noise model may produce highly diverse samples that have no relation to the training data.
Formally, we can view this in terms of the reverse Kullback--Leibler (KL) divergence~\citep{Papamakarios2019} of the generative model \(q\) with respect to the true distribution \(p\) on the sample space \(\mathcal{X}\):
\begin{equation}
    \mathcal{D}_\text{KL}(q \parallel p) = -H(q) - \E_q[\log p(X)],
    \label{eq:rev-kl}
\end{equation}
where \(H(q)\) denotes the differential entropy of \(q\) and \(\E_q\) is the expected value with respect to \(q\).
To achieve a good fit, that is, a low KL divergence, we thus want to simultaneously maximize the entropy \(H(q)\) and the log-likelihood \(\E_q[\log p(X)]\) of \(p\) under the approximation \(q\).

One popular family of generative models is that of the energy-based model (EBM)~\citep{Geman1984}, also known as a \textit{canonical} or \textit{macrocanonical ensemble}~\citep{Jaynes1957}, typically formulated as the Gibbs or Boltzmann distribution \(q(x) \propto \exp(-\beta \cdot \Phi(x))\) for a energy function \(\Phi: \mathcal{X} \to \R^K\) and parameter vector \(\beta \in \R^K\) (the inverse temperature).
This is the distribution that maximizes the entropy \(H(q)\) subject to the moment constraint \(\E_q[\Phi(X)] = \alpha\) for some target energy vector \(\alpha \in \R^K\)~\citep{Cover2006}.

In this work, we tackle the one-shot learning problem, where we are given \(\Phi\) and \(\alpha = \Phi(y)\) is obtained from some observation \(y \in \mathcal{X}\).
Here, \(\Phi\) may be given by some domain-specific design or earlier learning procedures.
Using the macrocanonical approach here suffers from two main challenges, namely determining \(\beta\) and sampling, both nontrivial in the general case and in particular when \(\mathcal{X}\) is high-dimensional.
As a remedy to the first issue is the \textit{microcanonical ensemble}~\citep{Lanford1975, Ellis2000, Touchette2015}, which is also a maximum-entropy distribution but constrained to distributions with support in the \textit{microcanonical set} of width \(\varepsilon > 0\),
\begin{equation}
    \Omega_\varepsilon \coloneqq \{x \in \mathcal{X}: \| \Phi(x) - \alpha \| \leq \varepsilon \}.
\end{equation}
Maximizing the entropy here implies that the distribution is uniform over this set. Thus, the entropy is equal to the log of the volume of \(\Omega_\varepsilon\) which is increasing in \(\varepsilon\).
This approximation relies on the assumption that \(\Phi(X)\) concentrates around its mean with high probability under the true distribution \(p\), which is the case for most stationary time series of sufficiently long duration and when \(\Phi\) is defined as the time average of time-shift equivariant potentials.
The parameter \(\varepsilon\) can then be adjusted to match this concentration of \(\Phi(X)\).

While the microcanonical ensemble avoids the issue of estimating \(\beta\) in the macrocanonical model, sampling in high-dimensional spaces remains challenging.
To mitigate this, the microcanonical gradient descent model~(MGDM) was introduced by \citet{Bruna2019} as an approximation of the microcanonical ensemble which is easier to sample from, and has been successfully applied in a variety of domains~\citep{Bruna2019, Leonarduzzi2019, Morel2023scatspectra, Zhang2021, Brochard2022, Cheng2023, Auclair2023}.
The MGDM is defined as the pushforward of Gaussian white noise by way of a sequence of gradient descent steps that seek to minimize the objective
\begin{equation}
    L(x) \coloneqq \frac{1}{2} \| \Phi(x) - \alpha\|^2. \label{eq:gd-loss}
\end{equation}
Thus, taking \(\mathcal{X} = \R^d\), samples from the MGDM are generated by sampling $x_0$ from $\mathcal{N}(0, \sigma^2\eye_d)$ for some initial variance $\sigma^2$ and updating the sample using
\begin{equation}
    x_{t+1} = g(x_t) \coloneqq x_t - \gamma J_\Phi^\T(x_t) (\Phi(x_t) - \alpha), \label{eq:gd-step}
\end{equation}
where \(\gamma\) is the step size and \(J_\Phi \in \R^{K \times d}\) is the Jacobian of \(\Phi\).
This is typically iterated for a fixed number of steps $T$ or until $x_t$ reaches the microcanonical set \(\Omega_\varepsilon\) for some fixed \(\varepsilon\)~\citep{Leonarduzzi2019, Morel2023scatspectra}.

Despite its success, MGDM can be shown to suffer from entropy collapse in many cases, resulting in a model that is able to produce typical samples but lacks sufficient variability.
We shall see that this is due to the contraction of the distribution that typically occurs with each gradient step, reducing the entropy and leading to a higher KL divergence.
To remedy this, we propose a new variant of the MGDM, called the mean-field microcanonical gradient descent model (MF--MGDM), which generates a batch of samples \(\bm{x} \coloneqq \{x^{(n)}\}_{n=1}^N\) such that their mean energy vector satisfies the necessary constraints, effectively replacing \(\Phi\) in \eqref{eq:gd-loss} with the batch mean 
\begin{equation}
    \overline{\Phi}(\bm{x}) \coloneqq \frac{1}{N} \sum\nolimits_{n=1}^N \Phi (x^{(n)}).
    \label{eq:mean-energy}
\end{equation}
In this model, the initial distribution is not so much contracted as transported through the energy space to the target while maintaining more of its initial entropy.
We provide a theoretical justification for this in the form of a tighter lower bound on the entropy.
The resulting model combines the expressiveness of the micro- and macrocanonical ensembles with the efficient sampling of the MGDM.
The choice of energy function \(\Phi\) is highly dependent on the particular distribution to be approximated.
To illustrate the power of the proposed approach, we therefore evaluate MF--MGDM for a range of possible functions.
In each case, we see a significant improvement over the basic MGDM approach, validating the theoretical results obtained on the entropy lower bound.

The structure of this article is as follows.
Section~\ref{sec:related-work} surveys the literature on energy-based models and the MGDM in particular, while Section~\ref{sec:overfitting} illustrates the entropy collapse observed in the MGDM.
A proposed solution to this is introduced in Section~\ref{sec:mf-mgdm} in the form of the MF--MGDM along with a lower bound on its entropy, and numerical results supporting this algorithm are presented in Section~\ref{sec:experiments}.
Python code to reproduce the results in this paper may be found at \url{https://github.com/MarcusHaggbom/mf-mgdm}.

\section{Related work}
\label{sec:related-work}
The micro- and macrocanonical ensembles are both maximum entropy distributions conditioned on the target energy \(\alpha\).
These are related via the Boltzmann equivalence principle~\citep{Lanford1975}, which states that under certain conditions of \(\Phi\), they converge to the same measure as \(\dim \mathcal{X} \to \infty\) and \(\varepsilon \to 0\).
While it is not guaranteed that a maximum entropy distribution exists in the macrocanonical case~\citep{Cover2006}, the microcanonical ensemble is more general in that it allows for a wider range of energy functions~\citep{Bruna2019}.
Both ensembles allow sampling by MCMC methods, which is computationally challenging, but have been employed in high dimensions for EBMs \citep{Du2019} and score-based diffusion models \citep{yang2023diffusionsurvey}. 
This relies on sufficient mixing of the Markov chain, which is crucial for obtaining reliable Monte Carlo estimates in finite time, e.g. of the expectations in \eqref{eq:rev-kl} when comparing models with respect to the reverse KL divergence.

The MGDM was introduced in \citet{Bruna2019} for the purpose of facilitating sampling. Each step is deterministic, allowing us to calculate the exact likelihood of each sample, which, unlike MCMC methods, makes computing entropy comparatively easy.
The MGDM has been used in a variety of applications, such as cosmology~\citep{Cheng2023, Auclair2023} and texture synthesis \citep{Brochard2022, Zhang2021}. In these contexts, the model is often paired with various extensions of the \textit{scattering transform}~\citep{Mallat_2012} used as features in the energy function. The scattering transform is a composition of wavelet transforms and non-linearities, and can be seen as a convolutional neural net with predefined weights~\citep{Mallat2016}. Apart from its use as an energy function in generative models, it has also found applications in image classification~\citep{Bruna2013, Oyallon2019}, audio similarity measurement~\citep{Anden2019, Lostanlen2021}, molecular energy regression~\citep{Eickenberg2017}, and heart beat classification~\citep{Warrick2020} among others.

In the context of finance, MGDMs coupled with variants of the scattering transform have been used to generate sample paths of time series. In \citet{Leonarduzzi2019}, it is shown that the time-average of the second-order scattering transform encodes heavy tails, and that including also phase harmonic correlations~\citep{Mallat2019} encapsulates temporal asymmetries, both of which are typical features of financial time series. An extension of this representation is the scattering spectrum~\citep{Morel2023scatspectra}, which increases sparsity and better captures multiscale properties of rough paths such as fractional Brownian motion.

Another popular feature representation for rough paths is the truncated \textit{signature} \citep{Lyons2014}. The full signature of a path is a lossless representation up to time parametrization, and the truncation error decreases as the inverse of the factorial of the number of included terms. Whereas the features based on the scattering transform are typically used as is, the truncated signature usually functions as a compact initial feature on top of which learning methods are applied. In financial time series generation, this encoding has proved efficient for other generative models, e.g. variational autoencoders~\citep{Buehler2020} and Wasserstein GANs~\citep{Ni2021}. In principle, these learned features could serve as energy function in the canonical ensembles.

\section{Overfitting to target energy}
\label{sec:overfitting}
With each gradient step, the MGDM pushes the energy vector \(\Phi(x)\) of a sample \(x\) from the initial distribution towards the target energy \(\alpha\).
In doing so, however, the distribution of \(x\) and \(\Phi(x)\) also contracts.
As a result, by the time the process reaches the microcanonical set \(\Omega_\varepsilon\), a significant reduction of entropy has been incurred, producing a poor fit to the microcanonical ensemble.

\subsection{An illustrative example}
As an example, we consider the AR(1) model with parameter \(\varphi\) and conditional variance \(\sigma^2\):
\begin{equation}
    x_{i} = \varphi x_{i-1} + \sigma \varepsilon_i,\label{eq:ar1-recursion}
\end{equation}
where \((\varepsilon_i)_i\) is Gaussian white noise. If \(|\varphi| < 1\), the process is stationary and has the marginal distribution \(x_i \sim \mathcal{N}(0, \sigma^2/(1-\varphi^2))\).
Assuming \(x_1\) is drawn from this marginal, the likelihood is
\begin{align*}
    p(x) &\propto \exp{\left\{ -\frac{1}{2\sigma^2}\sum_{i=2}^d (x_i - \varphi x_{i-1})^2 - \frac{1-\varphi^2}{2\sigma^2} x_1^2 \right\}} \approx \exp {\left\{ \frac{\varphi}{\sigma^2} \sum_{i=2}^d x_i x_{i-1} -\frac{1+\varphi^2}{2\sigma^2}\sum_{i=1}^d x_i^2 \right\}}.
\end{align*}
Thus, AR(1) is approximately an exponential family with the sufficient statistics
\begin{equation}
    \Phi(x) = \left(\frac{1}{d} \sum_{i=2}^d x_i x_{i-1} , \; \frac{1}{d} \sum_{i=1}^d x_i^2 \right),
    \label{eq:ar1-energy}
\end{equation}
and is by the Boltzmann equivalence principle asymptotically equivalent with the microcanonical approximation with energy function \eqref{eq:ar1-energy}.

\begin{figure}[t]
    \centering
    \subcaptionbox{MGDM, \(T=100\)\label{fig:overfitted-energy-ar1}}{\includegraphics{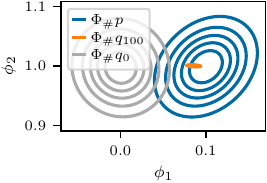}}
    \subcaptionbox{MGDM, \(T=36\)\label{fig:overfitted-energy-ar1-early}}{\includegraphics{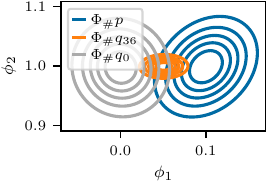}}
    \subcaptionbox{MF--MGDM, \(T=157\)\label{fig:ar1-energy-mf}}{\includegraphics{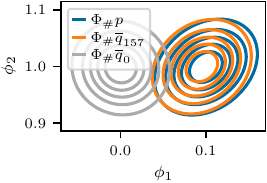}}
    \caption{Densities of \(\Phi(X)\), using fitted 2D Gaussians, at different stages of the descent for MGDM and MF--MGDM. In (b) and (c), \(T\) is the respective optimal number of steps to minimize KL divergence. The true distribution \(p\) is an AR(1) process with \(\varphi=0.1\) and \(\sigma^2 = 0.99\).}
    \label{fig:overfitted-ar1}
\end{figure}

Let us now approximate the microcanonical model using MGDM.
We thus have an initial measure \(q_0\) that is mapped through \(T\) steps of gradient descent to some final measure \(q_T\).
Figure~\ref{fig:overfitted-energy-ar1} illustrates how the initial distribution in the energy space \(\Phi_{\#} q_0\) is mapped to its final distribution \(\Phi_{\#} q_T\) after \(T = 100\) steps, bringing it close to the target energy. As can be seen in the pushforward of the true measure \(\Phi_{\#}p\), however, true samples have a much greater variability in these statistics, making clear the need for regularization. If we instead stop the gradient descent earlier, after \(T = 36\) steps, we obtain the distribution in Figure~\ref{fig:overfitted-energy-ar1-early}, where we have preserved more of the entropy, but at the cost of a worse likelihood fit. The MF--MGDM, which we introduce below, performs well with respect to both aspects (Fig.~\ref{fig:ar1-energy-mf}).

\begin{wrapfigure}{rt}{0.6\textwidth}
    \centering
    \vspace{-0.23in}
    \subcaptionbox{MGDM\label{fig:overfitted-kl-ar1}}{\includegraphics{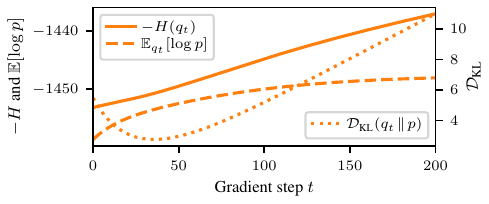}}
    \subcaptionbox{MF--MGDM\label{fig:ar1-kl-mf}}{\includegraphics{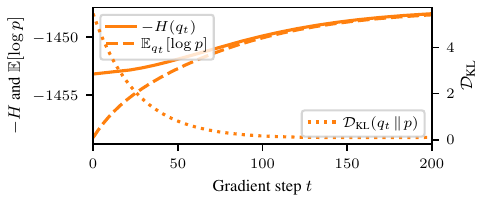}}
    \caption{Reverse KL divergence for the AR(1) example. The negative entropy and expected log-likelihood are plotted on the left-hand side, and the divergence on the right. \vspace{-0.59in}}
    \label{fig:kl-through-descent}
\end{wrapfigure}

\subsection{KL divergence}
Using the reverse KL divergence allows us to quantitatively analyze the method in examples like the AR(1) model where we have access to the density function of the target distribution. If \(\nabla L\) is Lipschitz and the step size \(\gamma\) is smaller than the Lipschitz constant, the gradient step \eqref{eq:gd-step} is contractive and MGDM can be seen as a contractive residual flow. The log-likelihood \(\log q_T\) is therefore
\begin{align}
    \label{eq:log-likelihood-flow}
    \begin{split}
        &\log q_T (x) = \log q_0 (z) \\ &\quad - \sum_{t=0}^{T-1} \log |\det J_{g}(G_t(z))|,
    \end{split}
\end{align}
where \(G_t\) denotes \(t\) compositions of \(g\) (with \(G_0 \coloneqq \eye\)), and \(z \coloneqq G_T^{-1}(x)\). The Jacobian \(\det J_g(G_t(z))\) is computed by automatic differentiation through \texttt{torch.func} in PyTorch~\citep[v2.1]{Paszke2019} (\href{https://opensource.org/license/BSD-3-Clause}{BSD-3}).
To arrive at the KL divergence, the expected values of \eqref{eq:log-likelihood-flow} and \(\log p\) in \eqref{eq:rev-kl} are estimated by Monte Carlo.

Going back to the AR(1) example, Figure~\ref{fig:overfitted-kl-ar1} illustrates how the reverse KL divergence attains its minimum after \(T=36\) steps and then starts increasing; the improvement in likelihood fit gradually diminishes while the entropy keeps decreasing, causing an entropy collapse. In this case, the trade-off between entropy and log-likelihood is a false trade-off in that minimizing the KL divergence leaves us with a poor entropy \emph{and} a poor expected log-likelihood, arguing against early stopping as a means of regularization.
In contrast, we see that the proposed method, MF--MGDM, does not exhibit this problem in Figure~\ref{fig:ar1-kl-mf}.

\section{Mean-field microcanonical gradient descent}
\label{sec:mf-mgdm}
In the MGDM, the expected log-likelihood increases as the descent progresses and the energy approaches the target (assuming an appropriate energy function for the given distribution we model, e.g. the sufficient statistics as in the AR(1) case). Conversely, if too many iterates are performed, the energy vectors of the samples will be too close.
Note that this happens even if the \(\varepsilon\) parameter of the microcanonical ensemble is chosen to be large, since the MGDM method will be concentrated over a small subset of \(\Omega_\varepsilon\). This observation leads to our proposition of the mean-field microcanonical gradient descent model (MF--MGDM).

\subsection{The model}

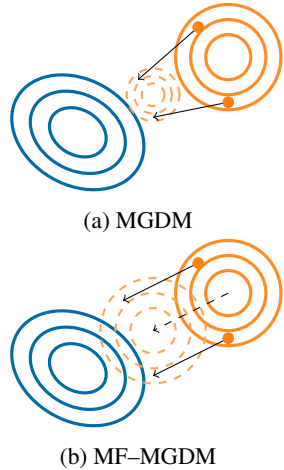
\begin{wrapfigure}{r}{0.35\textwidth}
    \centering
    \subcaptionbox{MGDM\label{fig:gradient-pull-regular}}{
        \begin{tikzpicture}
            \draw[rotate=60, color=myblue, very thick] (0,0) ellipse (0.3 and 0.4);
            \draw[rotate=60, color=myblue, very thick] (0,0) ellipse (0.5 and 0.67);
            \draw[rotate=60, color=myblue, very thick] (0,0) ellipse (0.7 and 0.93);
            
            \draw[color=myorange!90, very thick] (2,1) circle (0.3);
            \draw[color=myorange!90, very thick] (2,1) circle (0.5);
            \draw[color=myorange!90, very thick] (2,1) circle (0.7);

            \draw[color=myorange!70, thick, dashed] (1,0.5) circle (0.15);
            \draw[color=myorange!70, thick, dashed] (1,0.5) circle (0.25);
            \draw[color=myorange!70, thick, dashed] (1,0.5) circle (0.35);
            
            \draw[->] (2, 0.4) -- (1, 0.2);
            \filldraw [myorange] (2, 0.4) circle (2pt);
            \draw[->] (1.6, 1.4) -- (0.8, 0.7);
            \filldraw [myorange] (1.6, 1.4) circle (2pt);
        \end{tikzpicture}
    }
    \subcaptionbox{MF--MGDM\label{fig:gradient-pull-mf}}{
        \begin{tikzpicture}
            \draw[rotate=60, color=myblue, very thick] (0,0) ellipse (0.3 and 0.4);
            \draw[rotate=60, color=myblue, very thick] (0,0) ellipse (0.5 and 0.67);
            \draw[rotate=60, color=myblue, very thick] (0,0) ellipse (0.7 and 0.93);
            
            \draw[color=myorange!90, very thick] (2,1) circle (0.3);
            \draw[color=myorange!90, very thick] (2,1) circle (0.5);
            \draw[color=myorange!90, very thick] (2,1) circle (0.7);

            \draw[color=myorange!70, thick, dashed] (1,0.5) circle (0.3);
            \draw[color=myorange!70, thick, dashed] (1,0.5) circle (0.5);
            \draw[color=myorange!70, thick, dashed] (1,0.5) circle (0.7);

            \draw[->, dashed] (2, 1) -- (1, 0.5);
            \draw[->] (2, 0.4) -- (1, -0.1);
            \filldraw [myorange] (2, 0.4) circle (2pt);
            \draw[->] (1.6, 1.4) -- (0.6, 0.9);
            \filldraw [myorange] (1.6, 1.4) circle (2pt);
        \end{tikzpicture}
    }
    \caption{Illustration of \(\Phi\)-pushforward measures of the true distribution in blue centered close to the target energy \(\alpha\), and the approximation in orange. In the regular MGDM, each particle individually seeks to minimize its distance to the origin in energy space, potentially causing a collapse; in the mean-field version, the particles move approximately in parallel.}
    \vspace{-0.2in}
    \label{fig:gradient-pull}
\end{wrapfigure}

In the MF--MGDM, the mass of the initial distribution is pushed towards the target in energy space while attempting to reduce the collapse of the radius of the ball (or similarly the energy variance) and thereby reducing the entropy loss. The principle is illustrated in Figure~\ref{fig:gradient-pull}. Whereas the regular MGDM (Figure~\ref{fig:gradient-pull-regular}) updates each sample \emph{individually} with the objective of minimizing its energy distance \eqref{eq:gd-loss} to the target, MF--MGDM (Figure~\ref{fig:gradient-pull-mf}) updates several samples simultaneously so that they move towards the target energy \emph{in aggregate}.

Formally, define $\bm{x} = \{x^{(n)}\}_{n=1}^N \in \R^{Nd}$ as a collection of $N$ particles, where a \emph{particle} is a sample path in \(\R^d\). Recalling the mean energy \(\overline{\Phi}\) in \eqref{eq:mean-energy}, the new optimization objective is
\begin{equation}
    \overline{L}(\bm{x}) \coloneqq \frac{N}{2} \|\overline{\Phi}(\bm{x}) - \alpha\|^2.
    \label{eq:mf-loss}
\end{equation}
Denoting by \(\mathcal{J}_{\Phi}(\bm{x})\) the concatenation of the Jacobians \(J_\Phi(x^{(n)})\) of \(\Phi\) with respect to each particle \(x^{(n)}\),
\begin{equation}
    \mathcal{J}_{\Phi}(\bm{x}) \coloneqq \begin{bmatrix}
        J_\Phi(x^{(1)}) & \cdots & J_\Phi(x^{(N)})
    \end{bmatrix}  \in \R^{K \times Nd},
    \label{eq:batch-phi-jac}
\end{equation}
we define the mean-field gradient step as a gradient step for the objective \eqref{eq:mf-loss}, namely
\begin{align}
    \overline{g}(\bm{x}) \coloneqq \bm{x} - \gamma \mathcal{J}_\Phi^\T(\bm{x}) \left(\overline{\Phi}(\bm{x}) - \alpha\right).
    \label{eq:mf-step}
\end{align}

The mean-field concept originates from statistical physics as a tool for studying macroscopic phenomena in large particle systems by averaging over microscopic interactions. In the context of game theory, for instance, mean-field games are multiagent problems where each agent has a negligible impact on the others, so that the dynamics of an agent depends on the law of the system. For an \(N\)-player system, the law is the empirical measure, for which a subclass of systems are those where the dynamics depend on the empirical mean. The mean-field limit is then when \(N \to \infty\); see e.g. \citet{MF-games-book}. We can think of \eqref{eq:mf-step} as corresponding to a discretization of a system of differential equations with mean-field interactions.

The MF--MGDM faces two challenges that the regular model does not. The first is that the sampling procedure requires simultaneous generation of multiple samples in order to compute \(\overline{\Phi}\). This is solved efficiently by vectorizing the computation of \(\mathcal{J}_{\Phi}(\bm{x})\) in \eqref{eq:batch-phi-jac}. Furthermore, most applications call for generation of multiple samples, so the additional cost would be incurred at any rate by multiple invocations of MGDM.

The second challenge is that of computing the entropy, specifically computing the log-determinant of the Jacobian of a gradient step \(\overline{g}\). The issue is that the samples are now coupled, resulting in the Jacobian being one large \(Nd \times Nd\) matrix. Naively computing the determinant scales as \(\bigo(N^3d^3)\) (even keeping the Jacobian in memory is infeasible), but it is possible to rewrite it on a form that allows \(\bigo(Nd^3)\) computation by writing the Jacobian as a sum of a block diagonal and a low-rank matrix, and then using the matrix determinant lemma (see Appendix~\ref{section:appendix-computing-logdetjac}).

\subsection{An illustrative example -- revisited}

To demonstrate the effect of the mean-field gradient step, we return to the AR(1) example.
Figure~\ref{fig:ar1-energy-mf} shows the pushforward by $\Phi$ of the MF--MGDM approximation after 157 steps when minimum KL is achieved.
We see now that the final distribution in energy space more closely aligns with that of the true measure, preventing the reduction of entropy observed in the MGDM (see Figure~\ref{fig:overfitted-energy-ar1}).
Tracking the reverse KL divergence for each gradient step, Figure~\ref{fig:ar1-kl-mf}, we see an almost monotone decrease, avoiding the need for early stopping.
If we break up the KL divergence into negative entropy and log-likelihood, we no longer observe an unbounded decrease in entropy.
Instead, it stabilizes around a value close to the negative log-likelihood, resulting in a small KL divergence.

\begin{figure}[t]
\vskip 0.2in
\begin{center}
    \includegraphics{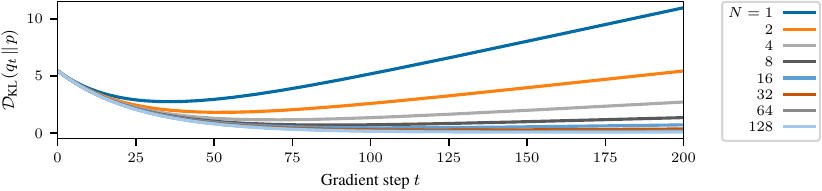}
    \caption{Reverse KL divergence through gradient descent with respect to the true model AR(1) for MF--MGDM with different mean-field batch sizes \(N\), and with Monte Carlo sample size 128.}
    \label{fig:ar1-mf-bs}
\end{center}
\vskip -0.2in
\end{figure}

\subsection{Theoretical entropy bound}
Since the entropy of \(d\) i.i.d. random variables scales linearly in \(d\) it is natural to define the \textit{entropy rate} \(d^{-1} H(p_1 \times \cdots \times p_d)\) for the joint distribution of sequences of random variables~\citep{Cover2006}. In MF--MGDM, the joint distribution is over \(N\) time series of length \(d\), hence we have to normalize with \(Nd\).
\begin{theorem}
    \label{thm:entropy-bound}
    Assume \(\Phi \in \bm{\mathrm{C}}^2\), with \(\beta\) and \(\eta\) denoting the Lipschitz constants of \(\Phi\) and \(\nabla \Phi\), respectively. Denote \(\overline{q}_T^N\) as the distribution of the MF--MGDM model with \(N\) particles after \(T\) iterations. Then the entropy rate \((Nd)^{-1}H(\overline{q}_T^N)\) admits, up to \(\bigo(\gamma^2)\) terms, the lower bound
    \[(Nd)^{-1}H(\overline{q}_T^N) \geq (Nd)^{-1}H(\overline{q}_0^N) - 2\gamma\left( \eta\sqrt{K}\sum_{t=0}^{T-1} \E_{\overline{q}_t^N}\|\overline{\Phi}(\bm{X}) - \alpha\| + \frac{K}{Nd} \beta^2 T \right).\]
\end{theorem}
The entropy bound for the regular MGDM is recovered when \(N=1\) (since \(\overline{\Phi}\) and \(\Phi\) are then equal). Herein lies an explanation for the improvement in KL of the MF--MGDM. In both models, \(\overline{\Phi}\) or \(\Phi\) goes to \(\alpha\), whereby the cost in entropy for each gradient step is after a point mainly driven by the \(\beta^2 T\)-term, which can be made arbitrarily small in MF--MGDM by increasing \(N\). This is also reflected empirically in Figure~\ref{fig:ar1-mf-bs} where a monotonic improvement of KL divergence is observed as \(N\) grows.
Note, however, that this is a lower bound, so it does not guarantee that MF--MGDM always preserves entropy better than MGDM (although this is observed numerically), but it does provide a better guarantee.
The proof of Theorem~\ref{thm:entropy-bound} is given in Appendix~\ref{section:entropy-proof}.

\section{Numerical experiments}
\label{sec:experiments}
To evaluate the performance of this sampling scheme, we apply it to both synthetic and real-world time series, where the latter is taken from applications in financial modeling.

\subsection{Synthetic data}\label{sec:synth-data}
To compare the different approximation models on synthetic data, we use time series models that have density functions in closed form, allowing for evaluation of the reverse KL divergence. We generate 10~000 samples of length 1~024 and take the average energy over these samples as target energy, to simulate the idealized setting where the true energy vector is known, avoiding bias. The KL divergence is estimated by generating 128 samples from the respective models and recording the divergence after each gradient step. 

\begin{table*}[t]
\caption{Minimum reverse KL divergence over \(T\) for different distributions and approximation models, where \textsc{Reg.} denotes the regular MGDM whereas \textsc{MF} is the mean-field version; \(N=128\).}
\label{tab:min-kl}
\centering
\begin{small}
\begin{sc}
\begin{tabular}{lrrrrrrrr}
\toprule
 & \multicolumn{2}{c}{ACF eqn.~\eqref{eq:ar1-energy}}& \multicolumn{2}{c}{ScatMean}& \multicolumn{2}{c}{ScatCov}& \multicolumn{2}{c}{ScatSpectra}\\
 & Reg.& MF& Reg.& MF& Reg.& MF& Reg.&MF\\
\midrule
AR\((0.1)\) & 2.76 & \textbf{0.09} & 4.24 & 1.99 & 5.47 & 4.04 & 5.44 & 2.32 \\
AR\((0.2, -0.1)\) & 9.44 & \textbf{3.81} & 17.98 & 10.55 & 25.91 & 14.84 & 27.33 & 9.60 \\
AR\((-0.1,  0.2,  0.1)\) & 30.04 & 26.39 & 20.98 & 15.18 & 29.55 & 18.01 & 28.46 & \textbf{10.13} \\
CIR\((1/2, 1, 1)\) & 219.40 & 214.65 & 170.99 & 168.88 & 121.17 & 59.21 & 105.05 & \textbf{30.78} \\
CIR\((1/\sqrt{2}, \sqrt{2}, 1)\) & 104.49 & \textbf{87.96} & 182.32 & 179.34 & 223.63 & 204.79 & 203.46 & 201.44 \\
\bottomrule
\end{tabular}
\end{sc}
\end{small}
\end{table*}

We used the following energy functions:
\begin{enumerate}[label=\roman*.]
    \item AR(1) approximate sufficient statistics \eqref{eq:ar1-energy} (or equivalently, autocovariance at lags 0 and 1);
    \item First moments of the second-order scattering transform (with complex modulus as nonlinearity), using filters from the Kymatio package \citep[v0.3]{kymatio} (\href{https://opensource.org/license/BSD-3-Clause}{BSD-3});
    \item Second moments of the second-order scattering transform, augmented with filters shifted by \(0\) and \(\pi/3\) in the first-order coefficients, and using ReLU of the real part as nonlinearity.
    Finally, we perform a dimensionality reduction by using principal component analysis (PCA) on transforms applied to Gaussian white noise;
    \item Scattering spectra from \citet{Morel2023scatspectra} (\href{https://opensource.org/license/mit}{MIT License}), taking the modulus of those coefficients which are complex, and thereby ignoring the phase. 
\end{enumerate}

These energy functions are applied to two types of synthetic data: autoregressive models of order $p$ ($\text{AR}(p)$) models and Cox--Ingersoll--Ross (CIR) models.

\paragraph{$\text{AR}(p)$}
An $\text{AR}(p)$ model with parameters \(\varphi_1, \dots, \varphi_p\) and \(\sigma\) is a generalization of the AR(1) process in \eqref{eq:ar1-recursion} and is defined by the recursion
\(x_i = \sum_{j=1}^p \varphi_j x_{i-j} + \sigma \varepsilon_i,\)
with white noise \((\varepsilon_i)_i\), and is stationary if the roots of the characteristic polynomial \(\pi(z) = 1 - \sum_j \varphi_j z^j\) are outside the unit circle. In Table~\ref{tab:min-kl}, the models are denoted AR\((\varphi_1, \dots, \varphi_p)\), and \(\sigma\) is chosen as to obtain unit marginal variance.

\paragraph{CIR}
The CIR model \citep{CIR1985} is a diffusion process that is commonly used for modelling short-term interest rates.
It is related to the Ornstein--Uhlenbeck process, which can be seen as a continuous version of AR(1), but differs in the way that the diffusion term is scaled by the square root of the rate \(r \in \R^+\) to give
\[d r_t = \kappa (\theta - r_t) dt + \sigma \sqrt{r_t} d W_t,\]
where \(W\) is a Brownian motion.
The process admits a stationary distribution, and the distribution at time \(t\) given the value at an earlier time \(s < t\) is a scaled noncentral \(\chi^2\) distribution which can be written in closed form, allowing for explicit evaluation of the likelihood of a discretization in an autoregressive fashion. The distribution of \(r_0\) can be taken to be the marginal distribution, i.e., a gamma distribution.
In the experiments, we use the discretization \(\Delta t = 1\), and the models are identified as CIR\((\kappa, \theta, \sigma)\).
The CIR process is non-negative, so projected gradient descent, described in Appendix~\ref{section:appendix-projected-gradient-descent}, has to be used when approximating this distribution in the context of MGDM.

\paragraph{Results}
For each model and energy function, the reverse KL divergence was computed at each step through the descent. The minimum divergence achieved is displayed in Table~\ref{tab:min-kl}. For every true distribution, we present results also for energy functions that are not necessarily a good choice, given the true model. We see here that MF--MGDM consistently outperforms MGDM.

\begin{figure}[t]
    \centering
    \subcaptionbox{AR\((0.2, -0.1)\)\label{fig:kl-div-ar2}}{\includegraphics{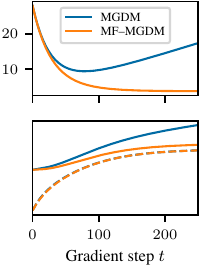}}
    \subcaptionbox{AR\((-0.1, 0.2, 0.1)\)\label{fig:kl-div-ar3}}{\includegraphics{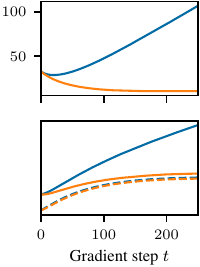}}
    \subcaptionbox{CIR\((1/2, 1, 1)\)\label{fig:kl-div-cir0}}{\includegraphics{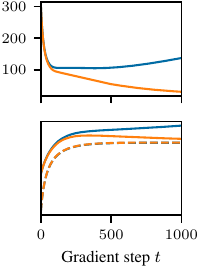}}
    \subcaptionbox{CIR\((1/\sqrt{2}, \sqrt{2}, 1)\)\label{fig:kl-div-cir1}}{\includegraphics{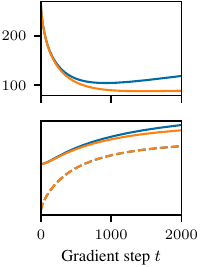}}
    \caption{Reverse KL divergence (top), negative entropy (bottom, solid) and log-likelihood (bottom, dashed) through the descent. Blue is regular MGDM and orange is MF--MGDM. The energy function used for each distribution is the corresponding optimal energy function according to Table~\ref{tab:min-kl}, i.e., (a) and (d) use ACF while (b) and (c) use scattering spectra. \(N=128\).}
    \label{fig:experiments-kl}
\end{figure}

The KL divergence through the descent as a function of iteration number is shown in Figure~\ref{fig:experiments-kl}, as well as its constituents entropy and expected log-likelihood. Here we have only plotted results for the energy function that best approximates each distribution in accordance with Table~\ref{tab:min-kl}. Here we see again that the mean-field model retains more entropy, and the difference is marginal between the expected likelihoods of the two models.

Another important difference here is that while MGDM needs to be stopped early to prevent the entropy from collapsing, this is not the case for MF--MGDM.
Indeed, we see that the entropy stabilizes after a certain number of steps similarly to the log-likelihood.
This is important because in a real-world setting, the true distribution is not known and the reverse KL divergence is not computable, so we cannot reasonably estimate the number of gradient steps to perform in order to balance the entropy loss with the increase of expected log-likelihood.
For MF--MGDM, we can run the sampling until convergence while being less sensitive to this type of overfitting.

\subsection{Financial data}
\begin{figure*}[hb]
    \centering
    \includegraphics{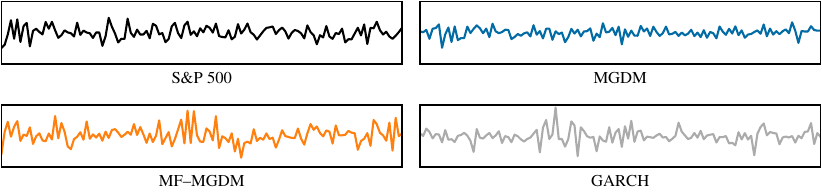}
    \caption{S\&P~500 realization and a randomly picked generated samples, showing a half-year window.}
    \label{fig:sp500-timeseries}
\end{figure*}

We evaluate the model on real financial data, namely the S\&P~500 index\footnote{Yahoo Finance \url{https://finance.yahoo.com/quote/\%5EGSPC/history} (\href{https://legal.yahoo.com/us/en/yahoo/terms/otos/index.html}{Terms})}, as well as five- and ten-year synthetic EUR and USD government bonds\footnote{Sveriges riksbank (Swedish Central Bank) \url{https://www.riksbank.se/en-gb/statistics/} (\href{https://www.riksbank.se/en-gb/statistics/interest-rates-and-exchange-rates/explanations--interest-rates-and-exchange-rates/frequently-asked-questions-about-exchange-rates/}{Terms})} quoted in yield.
For the equity index, daily log-returns are generated, while regular daily returns are used for the rates.
We use \(2^{12}\) points (\(\sim\) 16 years of daily data) for S\&P~500 and USD rates, and \(2^{11}\) (\(\sim\) 8 years) for EUR rates, which we divide into four samples of equal length.
The energy used for generation is estimated on the first sample, and the remaining three are used for validation.

\begin{figure*}[t]
    \centering
    \includegraphics{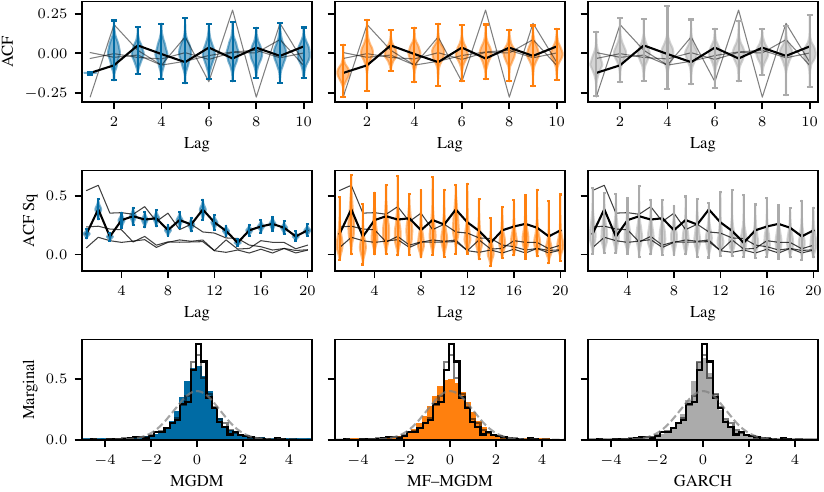}
    \caption{S\&P~500 autocorrelation, autocorrelation of the square of the signal, and marginal histogram. The top two rows are violin plots illustrating the empirical marginal density of the statistics for the models, with whiskers indicating min and max. In the top two rows, the statistics of the true sample used for the target energy \(\alpha\) is in black and the validation samples are thin gray. In the bottom row, the same holds, except the joint histogram for the validation data is shown. The dashed gray line is a Gaussian with moments matched to the true data. Generation time was equal for MGDM and MF--MGDM.}
    \label{fig:sp500}
\end{figure*}

As energy function \(\Phi\) we use statistics of interest for financial time series, namely variance, autocovariance at lag 1, and autocovariance of the squared process for lags 1--20.
(In Appendix~\ref{section:appendix-more-plots}, we also provide results for the scattering covariance as energy.)
Realizations of the models conditioned on S\&P~500 data are displayed in Figure~\ref{fig:sp500-timeseries} together with a slice of the validation data.
As reference, we also include a GARCH\((1, 1)\) model with AR\((1)\) mean process and Student's~t innovations, using maximum likelihood parameters fitted using the Python ARCH package \citep[v6.2]{archpackage} (\href{https://opensource.org/license/uoi-ncsa-php}{University of Illinois/NCSA Open Source License}).
All three models evaluated provide samples that are qualitatively similar to the original signal.
In addition, we compare the statistics included in \(\Phi\) and the marginal histograms, with results displayed in Figure~\ref{fig:sp500} (S\&P~500) and Figure~\ref{fig:rates-results} in Appendix~\ref{section:appendix-more-plots} (rates data).
The same general behavior is observed here as for the AR\((1)\) example, namely that the MF--MGDM counteracts the overfitting while still producing a good fit of the statistics that the model is conditioned on, comparable to GARCH.
The marginal fit, however, is superior for GARCH, with MF--MGDM becoming slightly worse than MGDM.
As a remedy, the energy function could be extended to include more sophisticated statistics to incorporate the heavy tails in both gradient models.
Finally, we estimated the entropy for the two microcanonical approximations to --48~800 for MGDM and +1~200 for MF--MGDM, in relation to +1~450 for Gaussian white noise, indicating improved performance with the mean-field approach.

\section{Limitations}
\label{sec:limitations}
First, we emphasize the stationarity assumption of the time series. Next, the MGDM requires the energy function \(\Phi\) to be differentiable so it is not straightforward to include e.g. order statistics constraints. As far as we know, there is presently no modification of the MGDM which allows for a stable way of inverting the descent in order to be able to compute forward KL in the usual case where the true distribution is not known, forcing only qualitative evaluation of performance on real-world data. Note also that in this case, any (differentiable) evaluation statistic that is of interest can be included in \(\Phi\), which in turn risks weakening the merit of the evaluation akin to Goodhart's law.  Finally, although the width \(\varepsilon\) of \(\Omega_\varepsilon\) is important for a good KL fit, exactly how to tune this parameter is left for future work.

\section{Conclusions}
\label{sec:conclusions}
The MGDM provides efficient sampling of high-dimensional distributions, but can suffer from a significant loss of entropy. Propagating too far into the descent is shown to overfit to the target energy that the model is conditioned on, meaning that the variance of the energy for the model is much too small as for what to expect from true distributions. Regularizing by early stopping in the descent mitigates this issue somewhat, but at the price of a worse fit to the true distribution and a larger bias from the initial distribution. The mean-field regularization of the model in the form of MF--MGDM leverages parallel sampling to mitigate the problem, improving the rate at which entropy is lost without a significant impact on the likelihood fit.
Future work will explore better initial distributions and more sophisticated update steps.
These will in turn open the door to considering forward KL divergence metrics, removing the need for access to the likelihood of the target distribution.

\begin{ack}
This work was partially supported by the Wallenberg AI, Autonomous Systems and 
Software Program (WASP) funded by the Knut and Alice Wallenberg Foundation.

The computations were enabled by the Berzelius resource provided by the Knut and Alice Wallenberg Foundation at the National Supercomputer Centre.

\end{ack}

\bibliography{WASP, refs}
\bibliographystyle{unsrtnat}

\newpage
\appendix
\section{Appendix -- Computing the Jacobian determinant in MF--MGDM}
\label{section:appendix-computing-logdetjac}
Without loss of generality, assume \(\alpha = 0\) (otherwise, we simply redefine \(\Phi(x)\) to be \(\Phi(x) - \alpha\)). Denote \(\overline{g}^{(n)}\) as the update corresponding to particle \(x^{(n)}\):
\begin{align*}
    \overline{g}^{(n)}(\bm{x}) = x^{(n)} - \gamma \sum_{k=1}^K \nabla \Phi_k(x^{(n)}) \overline{\Phi}_k(\bm{x}).
\end{align*}
Then the Jacobian w.r.t. a possibly different particle \(x^{(m)}\) is, stated by index,
\begin{align*}
    \partial_{x^{(m)}_j} \overline{g}^{(n)}_i (\bm{x}) &= \kronecker_{m, n}\kronecker_{i, j} - \gamma \sum_k \partial_{x^{(m)}_j} \left(\partial_{x^{(n)}_i} \Phi_k(x^{(n)}) \cdot \overline{\Phi}_k(\bm{x}) \right) \\
    &= \kronecker_{m, n}\kronecker_{i, j} - \gamma \sum_k \left(\kronecker_{m, n} \partial_{x^{(n)}_j} \partial_{x^{(n)}_i} \Phi_k(x^{(n)}) \cdot \overline{\Phi}_k(\bm{x}) + \frac{1}{N} \partial_{x^{(n)}_i} \Phi_k(x^{(n)}) \cdot \partial_{x^{(m)}_j} \Phi_k (x^{(m)}) \right),
\end{align*}
or, stated by block,
\begin{align*}
    J_{\overline{g}^{(n)}}(x^{(m)}) = \kronecker_{m, n} \cdot \left( \eye_d - \gamma \sum_k H_{\Phi_k}(x^{(n)}) \overline{\Phi}_k(\bm{x}) \right) - \frac{\gamma}{N} J_\Phi^\T(x^{(n)}) J_\Phi(x^{(m)}),
\end{align*}
where \(\eye_d\) is the \(d \times d\) identity matrix. Recall the concatenation \eqref{eq:batch-phi-jac} of the Jacobians,
\begin{align*}
    \mathcal{J}_\Phi (\bm{x}) = \begin{bmatrix}
        J_\Phi(x^{(1)}) & \cdots & J_\Phi(x^{(N)})
    \end{bmatrix},
\end{align*}
and define the block-diagonal matrix
\begin{align}
    \mathcal{H}_{\Phi_k} (\bm{x}) = \diag \left\{H_{\Phi_k}(x^{(n)})\right\}_{n = 1}^N 
    = \begin{bmatrix}
        H_{\Phi_k}(x^{(1)}) & & \\
        & \ddots & \\
        & & H_{\Phi_k}(x^{(N)})
    \end{bmatrix}.
    \label{eq:batch-phi-hess}
\end{align}
Then, the entire Jacobian of \(\overline{g}\) can be expressed as 
\begin{align}
    J_{\overline{g}} (\bm{x}) = \eye_{Nd} - \gamma \sum_k \mathcal{H}_{\Phi_k} (\bm{x}) \overline{\Phi}_k(\bm{x}) - \frac{\gamma}{N} \mathcal{J}_\Phi^\T (\bm{x}) \mathcal{J}_\Phi (\bm{x}).
    \label{eq:jac-gradstep-mf}
\end{align}
Using the matrix determinant lemma, and that
\[\eye_{Nd} - \gamma \sum_k \mathcal{H}_{\Phi_k} \overline{\Phi}_k\]
is block-diagonal (and thereby also its inverse), the determinant can be reformulated as
\begin{align*}
    \det J_{\overline{g}} &= \det \left( \eye_{Nd} - \gamma \sum_k \mathcal{H}_{\Phi_k} \overline{\Phi}_k - \frac{\gamma}{N} \mathcal{J}_\Phi^\T \mathcal{J}_\Phi\right) \\
    &= \det \left( \eye_{Nd} - \gamma \sum_k \mathcal{H}_{\Phi_k} \overline{\Phi}_k\right) \det \left( \eye_K  - \frac{\gamma}{N} \mathcal{J}_\Phi \left( 
    \eye_{Nd} - \gamma \sum_k \mathcal{H}_{\Phi_k} \overline{\Phi}_k \right)^{-1} \mathcal{J}_\Phi^\T \right) \\
    &= \det \diag \left\{\left(\eye_d - \gamma \sum_k H_{\Phi_k}^{(n)} \overline{\Phi}_k \right)\right\}_n \det \left( \eye_K  - \frac{\gamma}{N} \mathcal{J}_\Phi \diag \left\{\left(\eye_d - \gamma \sum_k H_{\Phi_k}^{(n)} \overline{\Phi}_k \right)^{-1} \right\}_n \mathcal{J}_\Phi^\T \right) \\
    &= \prod_n \det \left(\eye_d - \gamma \sum_k H_{\Phi_k}^{(n)} \overline{\Phi}_k \right) \det \left( \eye_K - \gamma \frac{1}{N} \sum_n J_\Phi^{(n)} \left(\eye_d - \gamma \sum_k H_{\Phi_k}^{(n)} \overline{\Phi}_k \right)^{-1} \left(J_\Phi^{(n)}\right)^\T\right).
\end{align*}

\section{Appendix -- Proof of Theorem~\ref{thm:entropy-bound}}
\label{section:entropy-proof}

As in previous Section~\ref{section:appendix-computing-logdetjac}, we assume without loss of generality that \(\alpha = 0\).

From \eqref{eq:log-likelihood-flow} we get
\begin{align}
    H(\overline{q}_T^N) = -\E_{\overline{q}_T^N}[\log \overline{q}_T^N (\bm{X})] &= -\E_{\overline{q}_0^N}\left[\log \overline{q}_0^N (\bm{X}) - \sum_{t=0}^{T-1}\log |\det J_{\overline{g}}(\overline{g}_t(\bm{X}))|\right] \nonumber \\
    &= H(\overline{q}_0^N) + \sum_{t=0}^{T-1}\E_{\overline{q}_t^N}[\log |\det J_{\overline{g}} (\bm{X})|.]
    \label{eq:entropy-rate-qtilde}
\end{align}
so we want to lower-bound \(\log|\det J_{\overline{g}}|\). By \eqref{eq:jac-gradstep-mf} we see that we can write \(J_{\overline{g}}(\bm{x})\) on the form \(\eye - \gamma A\). We have
\[\left. \frac{d}{d\gamma} \det (\eye - \gamma A) \right|_{\gamma=0} = -\det(\eye)\tr (\eye^{-1} A) = -\tr A, \]
which yields the Taylor approximation
\[\det (\eye - \gamma A) = 1 - \gamma \tr A + \bigo(\gamma^2).\]
This, together with the lower bound for the logarithm
\[\log(1-x) \geq -2x \]
for \(x \in [0, \frac{3}{4}]\),
results in the lower bound (suppressing the argument \((\bm{x})\))
\begin{equation}
    \log|\det J_{\overline{g}}| \geq -2 \gamma \left|\tr \left( \sum_k \mathcal{H}_{\Phi_k} \overline{\Phi}_k + \frac{1}{N} \mathcal{J}_\Phi^\T \mathcal{J}_\Phi \right)\right| + \bigo(\gamma^2)
    \label{eq:log-det-J-bound-mf}
\end{equation}
for \(\gamma\) small enough. Thus, we seek an upper bound to 
\begin{equation}
    \left|\tr \left( \sum_k \mathcal{H}_{\Phi_k} \overline{\Phi}_k + \frac{1}{N} \mathcal{J}_\Phi^\T \mathcal{J}_\Phi \right)\right| \leq \sum_k |\tr (\mathcal{H}_{\Phi_k}) \overline{\Phi}_k| + \frac{1}{N} |\tr( \mathcal{J}_\Phi^\T \mathcal{J}_\Phi)|.
    \label{eq:trace-bound-mf}
\end{equation}

The Lipschitz assumption on \(\Phi\) yields \(\|J_\Phi(x)\|_2 \leq \beta\) for all \(x\), so that for any particle (here suppressing the argument \((x^{(i)})\)),
\[\tr ( J_{\Phi}^\T J_{\Phi}) = \tr ( J_{\Phi} J_{\Phi}^\T) = \sum_k \lambda_k(J_{\Phi} J_{\Phi}^\T) \leq K \lambda_{\max}(J_{\Phi} J_{\Phi}^\T) = K\|J_{\Phi}^\T\|_2^2 = K\|J_{\Phi}\|_2^2 \leq K \beta^2,\]
whereby the second term of \eqref{eq:trace-bound-mf} becomes
\begin{equation}
    \frac{1}{N} \tr ( \mathcal{J}_{\Phi}^\T (\bm{x}) \mathcal{J}_{\Phi} (\bm{x})) = \frac{1}{N} \sum_{i=1}^N \tr (J_\Phi^\T (x^{(i)}) J_\Phi (x^{(i)})) \leq \frac{1}{N} N K \beta^2 = K \beta^2.
    \label{eq:trace-bound-jac-part}
\end{equation}

Similarly, the Lipschitz assumption on \(\nabla \Phi\) together with symmetry of \(H\) implies \(\|H_{\Phi_k}(x)\|_2 = |\lambda|_{\max}(H_{\Phi_k}(x)) \leq \eta\) for all \(k\) and \(x\), and in turn,
\[\sum_k | \tr ( H_{\Phi_k} ) \overline{\Phi}_k | \leq \sum_k d|\lambda|_{\max} (H_{\Phi_k}) |\overline{\Phi}_k| \leq d \eta \|\overline{\Phi}\|_1 \leq d \eta \sqrt{K} \|\overline{\Phi}\|_2.\]

Thus, the first term of \eqref{eq:trace-bound-mf} becomes
\begin{equation}
    \sum_k |\tr (\mathcal{H}_{\Phi_k} (\bm{x})) \overline{\Phi}_k (\bm{x})| = \sum_k \left|\sum_{i=1}^N \tr (H_{\Phi_k} (x^{(i)})) \overline{\Phi}_k (\bm{x})\right| \leq Nd\eta \sqrt{K}\|\overline{\Phi} (\bm{x})\|_2.
    \label{eq:trace-bound-hess-part}
\end{equation}

Inserting \eqref{eq:trace-bound-jac-part} and \eqref{eq:trace-bound-hess-part} into \eqref{eq:trace-bound-mf}, we see that the \(\log|\det J_{\overline{g}}|\) bound \eqref{eq:log-det-J-bound-mf} becomes
\begin{equation*}
    \log|\det J_{\overline{g}} (\bm{x})| \geq -2 \gamma \left( Nd \eta \sqrt{K}\|\overline{\Phi} (\bm{x})\|_2 + K \beta^2 \right) + \bigo(\gamma^2).
\end{equation*}

Hence, the lower bound on the entropy rate, up to second order terms in \(\gamma\), becomes
\begin{align*}
    (Nd)^{-1} H(\overline{q}_T^N) &= (Nd)^{-1} H(\overline{q}_0^N) - 2\gamma\left( \eta\sqrt{K}\sum_{t=0}^{T-1} \E_{\overline{q}_t^N}\|\overline{\Phi}(\bm{X})\|_2 + \frac{K}{Nd} \beta^2 T \right).
\end{align*}

\section{Appendix -- Projected gradient descent}
\label{section:appendix-projected-gradient-descent}
In the projected gradient descent used for CIR models, the generating procedure is to update the sample according to the gradient steps while satisfying the constraint of remaining in the positive cone \(x \geq 0\). A basic implementation is to alternate between a gradient step and a projection step, where the updated sample is projected onto the feasible set, which in practice amounts to applying a ReLU to the sample after each step; let \(g: \mathcal{X} \to \mathcal{X}\) denote the gradient update (regular or mean-field) and \(\underline{g}\) the projected gradient update, then
\[\underline{g} = \relu{} \circ g.\]
The problem with this definition is that the Jacobian becomes singular if an update is masked by the ReLU, resulting in the determinant being zero. Therefore, we instead use the update 
\[\underline{g}_i(x) = \begin{cases}
    g_i(x), & g_i(x) \geq 0, \\
    x_i, & g_i(x) < 0.
\end{cases}\]
Hence, if a component in the sample is negative after the gradient step \(g\), it is replaced by its prior value. In this case, the Jacobian determinant is the same as only looking at the components of the sample that have been updated.

Another aspect of the projected version of MGDM is the choice of initial measure. If the support of the marginal distribution is all of \(\R\), the maximum entropy distribution conditioned on the first two moments is the Gaussian. Thus, in this case, the MGDM is initialized with Gaussian white noise. For the CIR process, the support of the marginal distribution is \(\R^+\), and, given that it exists, the corresponding maximum entropy distribution is either the exponential (if the mean and standard deviation are equal) or the truncated Gaussian~\citep{Dowson1973}.

\section{Appendix -- Additional plots}
\label{section:appendix-more-plots}

\begin{figure}[h]
    \centering
    \subcaptionbox{EUR5Y\label{fig:eur5y}}{\includegraphics[width=0.49\textwidth]{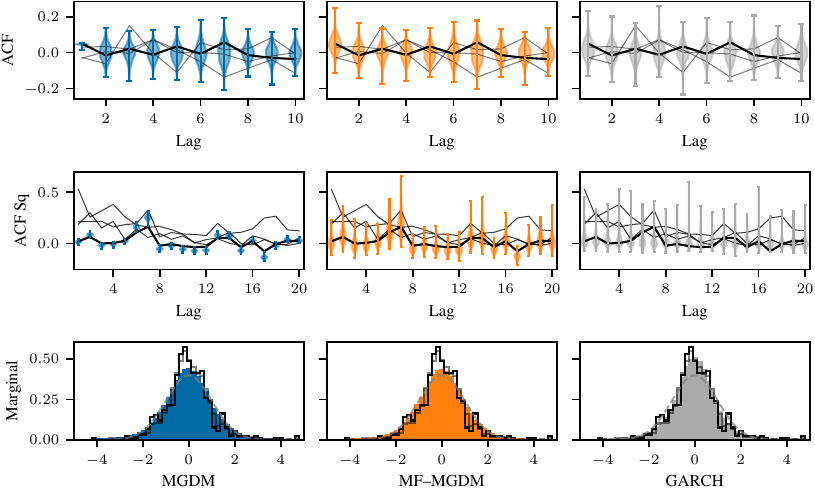}}
    \subcaptionbox{EUR10Y\label{fig:eur10y}}{\includegraphics[width=0.49\textwidth]{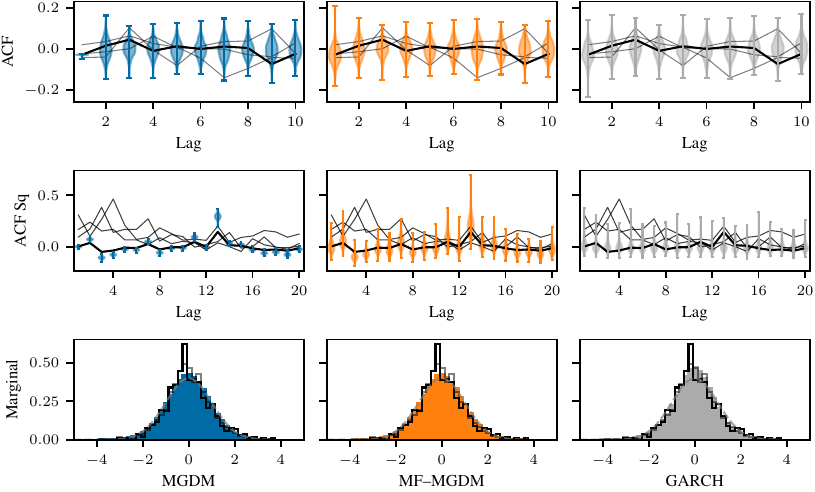}}
    \subcaptionbox{USD5Y\label{fig:usd5y}}{\includegraphics[width=0.49\textwidth]{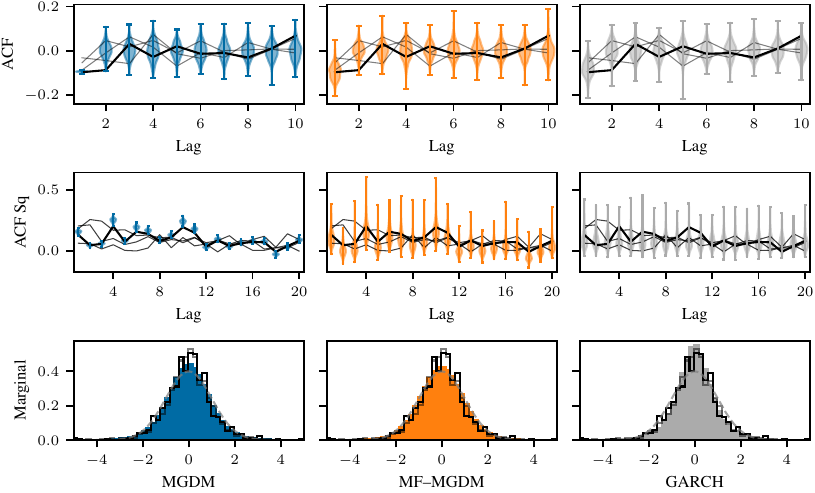}}
    \subcaptionbox{USD10Y\label{fig:usd10y}}{\includegraphics[width=0.49\textwidth]{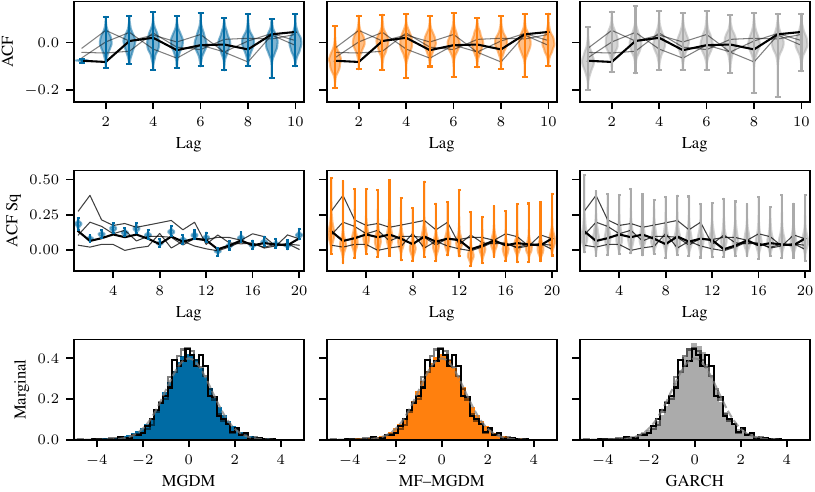}}
    \caption{Autocorrelations and marginal histograms as in Figure~\ref{fig:sp500}, with same energy function as for S\&P 500.}
    \label{fig:rates-results}
\end{figure}

\begin{figure}[h]
    \centering
    \subcaptionbox{S\&P 500\label{fig:sp500-scov}}{\includegraphics[width=0.98\textwidth]{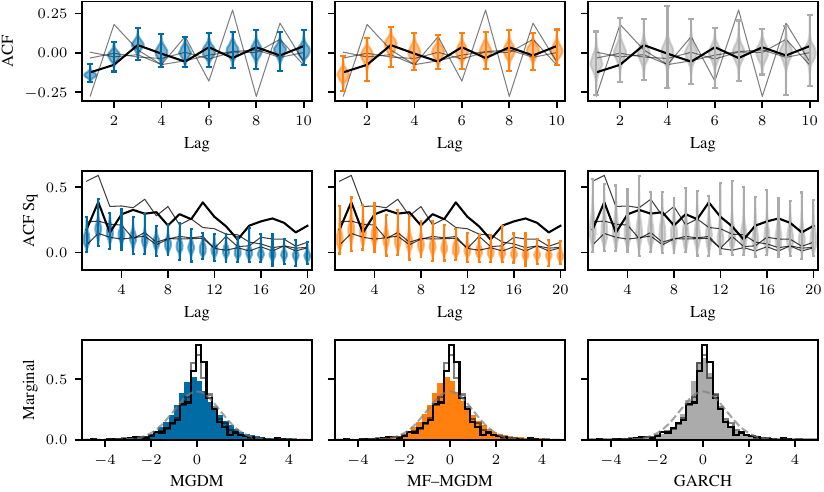}}
    \subcaptionbox{EUR5Y\label{fig:eur5y-scov}}{\includegraphics[width=0.49\textwidth]{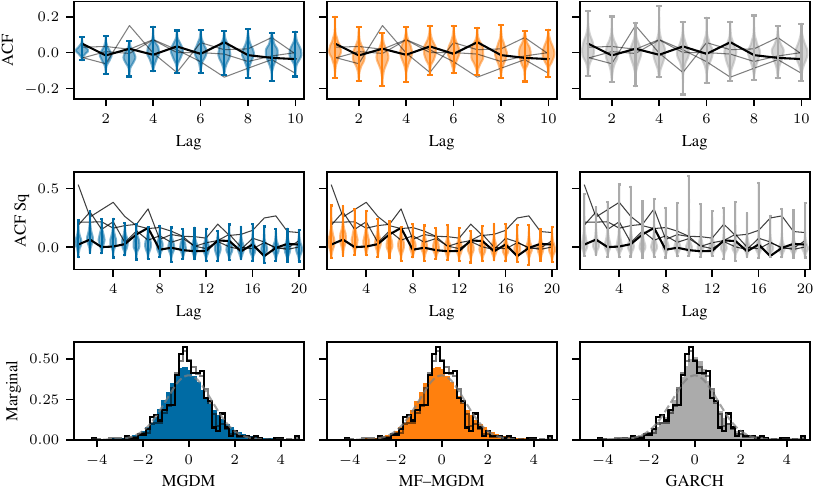}}
    \subcaptionbox{EUR10Y\label{fig:eur10y-scov}}{\includegraphics[width=0.49\textwidth]{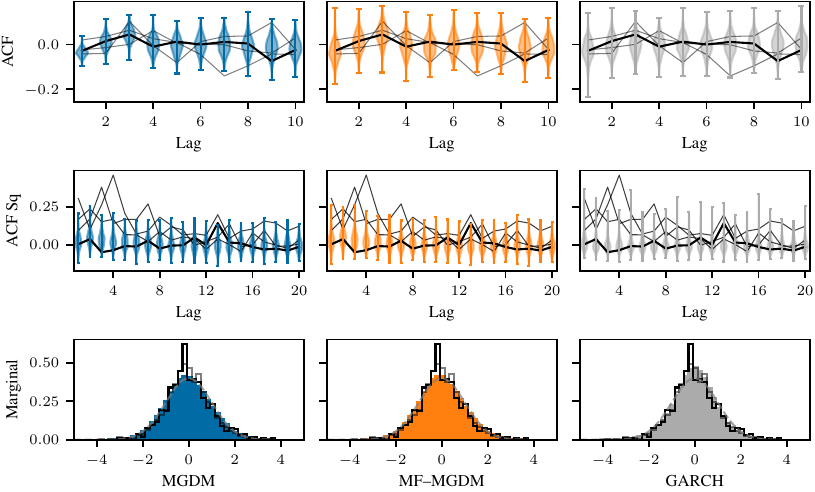}}
    \subcaptionbox{USD5Y\label{fig:usd5y-scov}}{\includegraphics[width=0.49\textwidth]{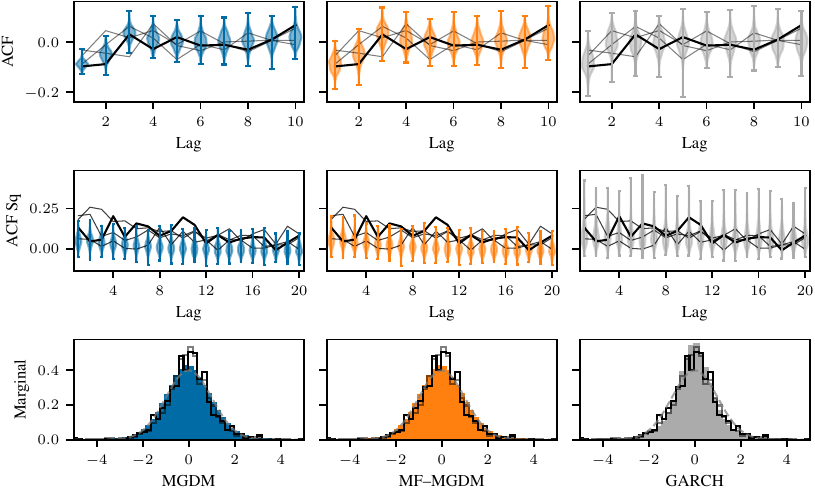}}
    \subcaptionbox{USD10Y\label{fig:usd10y-scov}}{\includegraphics[width=0.49\textwidth]{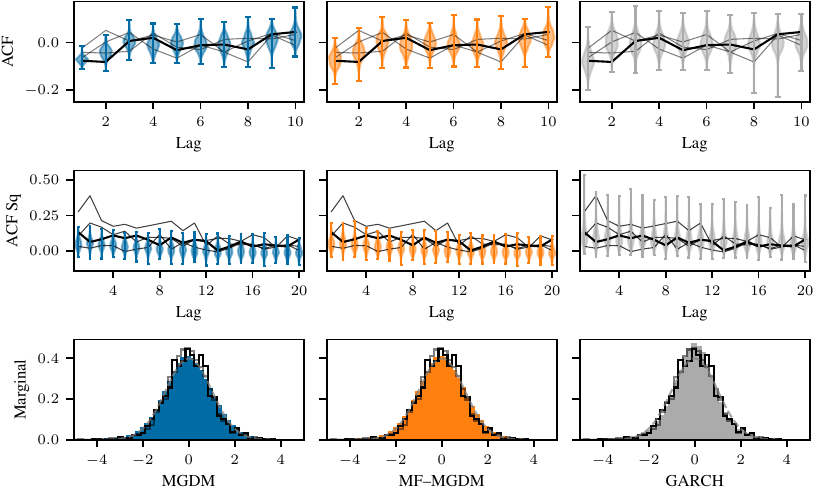}}
    \caption{Autocorrelations and marginal histograms as in Figure~\ref{fig:sp500}, here using the scattering covariance with phase shifts as described in Section~\ref{sec:synth-data}, but with PCA components now computed using samples from a GARCH process. Since the autocorrelations are not explicitly included in the energy function, the fit is worse. For MGDM, the statistics are not as concentrated as in Figures~\ref{fig:sp500} and \ref{fig:rates-results}. The MF--MGDM still provide \textit{some} improvements, most noticeable in the ACF.}
    \label{fig:rates-results-scov}
    \vskip -0.2in
\end{figure}

\newpage

\begin{figure}[h]
    \vskip 0.2in
    \centering
    \includegraphics[width=0.65\textwidth]{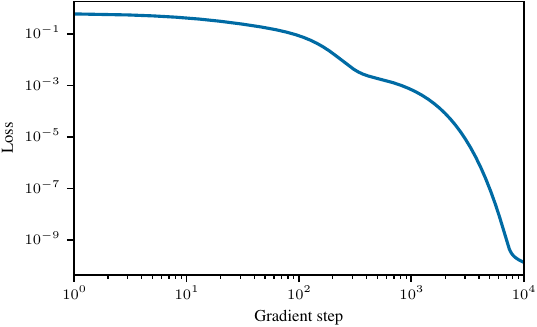}
    \caption{S\&P 500 average loss \eqref{eq:gd-loss} through the descent for the MGDM. The loss can be made arbitrarily small with more descent iterations, implying that the discrepancy in fit of the ACF of the squared signal in Figure~\ref{fig:sp500-scov} is not due to getting stuck in a poor local minimum.}
    \label{fig:sp500-loss-though-descent}
    \vskip -0.2in
\end{figure}

\begin{figure}[h]
    \vskip 0.2in
    \centering
    \includegraphics[width=0.98\textwidth]{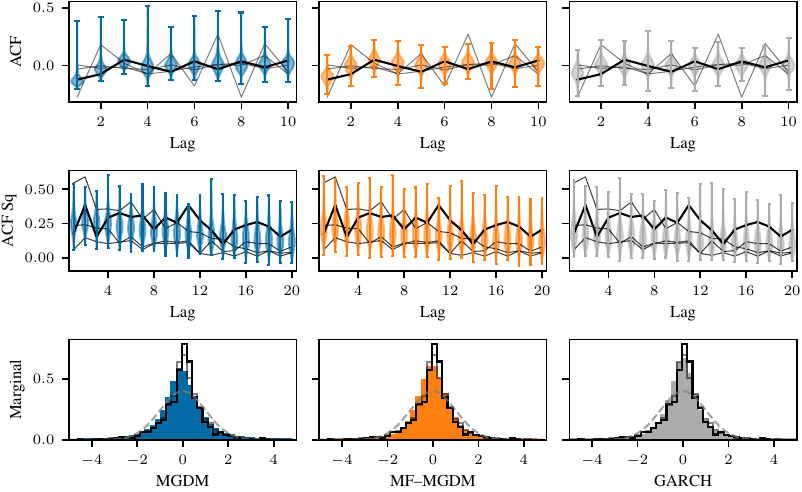}
    \caption{Autocorrelations and marginal histograms with scattering covariance energy as in Figure~\ref{fig:rates-results-scov}, the initial distribution now coming from a GARCH process as opposed to Gaussian white noise in Figure~\ref{fig:sp500-scov}. The fit is better for both MGDM and MF--MGDM compared to Fig.~\ref{fig:sp500-scov}. Together with Figure~\ref{fig:sp500-loss-though-descent}, this shows that the shortcomings in Fig.~\ref{fig:sp500-scov} are due to the microcanonical set being too large (i.e., additional moment constraints are necessary), rather than issues with the descent failing to transport the initial samples to the microcanonical set.}
    \label{fig:sp500-init-garch}
    \vskip -0.2in
\end{figure}

\end{document}